\documentclass[sigplan]{acmart}

\renewcommand\footnotetextcopyrightpermission[1]{} % removes footnote with conference information in first column
\usepackage{siunitx}            % typset units correctly
\usepackage{courier}            % standard fixed width font
\usepackage{comment}
\usepackage[scaled]{helvet} % see www.ctan.org/get/macros/latex/required/psnfss/psnfss2e.pdf
\usepackage{url}                  % format URLs
\usepackage{listings}          % format code
\usepackage{enumitem}      % adjust spacing in enums
\usepackage{xspace}
\usepackage{lipsum}
\usepackage{cleveref}
\usepackage{subcaption}
\usepackage{soul}
\usepackage[frozencache=true,cachedir=minted-cache]{minted} 
\definecolor{bg}{rgb}{0.5,0.5,0.5}
\definecolor{bggray}{rgb}{0.9, 0.9, 0.9}
\definecolor{brickred}{rgb}{0.8, 0.25, 0.33}
\definecolor{brightpink}{rgb}{1.0, 0.0, 0.5}
\definecolor{bluegray}{rgb}{0.4, 0.6, 0.8}

\newcommand{\sys}{KEN\xspace}

\newcommand{\ev}{eBPFNLDataset\xspace}
\newcommand{\compdataset}{KernelCompDataset\xspace}

\newcommand{\systp}{80\%}
\newcommand{\basetp}{30\%}
\newcommand{\sysimprovetp}{2.67\xspace}

\newcommand{\grumbler}[3]{\noindent{\color{#1}{\bf \fbox{#2}} {\it #3}}}
\newcommand{\arq}[1]{\grumbler{red}{ARQ}{#1}}
\newcommand{\yy}[1]{\grumbler{cyan}{YY}{#1}}

\begin{document}

\title{\sys: Kernel Extensions using Natural Language}

%
% any author declaration will be ignored  when using 'pldi' option (for double blind review)
%

\author{Yusheng Zheng}
\affiliation{%
  \institution{eunomia-bpf Community}
  \city{Shanghai}
  \state{Shanghai}
  \postcode{200000}
  \country{China}
}
\email{yunwei356@gmail.com}

\author{Yiwei Yang}
\affiliation{%
  \institution{UC Santa Cruz}
  \city{Santa Cruz}
  \state{California}
  \postcode{95060}
  \country{USA}
}
\email{yyang363@ucsc.edu}

\author{Maolin Chen}
\affiliation{%
  \institution{eunomia-bpf Community}
  \city{Shanghai}
  \state{Shanghai}
  \postcode{200000}
  \country{China}
}
\email{agaaain.try@gmail.com}

\author{Andrew Quinn}
\affiliation{%
  \institution{UC Santa Cruz}
  \city{Santa Cruz}
  \state{California}
  \postcode{95060}
  \country{USA}
}
\email{aquinn@ucsc.edu}

% \author{Andrew Quinn}
% \affiliation{%
%   \institution{University of California, Santa Cruz}
%   \streetaddress{Computer Science Engineering}
%   \city{Santa Cruz}
%   \state{California}
%   \postcode{95060}

% }
% \email{aquinn1@ucsc.edu}

\begin{abstract}
The ability to modify and extend an operating system is an important feature for improving a system's security, reliability, and performance.  The extended Berkeley Packet Filters (eBPF) ecosystem has emerged as the standard mechanism for extending the Linux kernel and has recently been ported to Windows.  eBPF programs inject new logic into the kernel that the system will execute before or after existing logic.  While the eBPF ecosystem provides a flexibility mechanism for kernel extension, it is difficult for developers to write eBPF programs today.  An eBPF developer must have deep knowledge of the internals of the operating system to determine where to place logic and cope with programming limitations on the control flow and data accesses of their eBPF program enforced by the eBPF verifier.  In the end, the limitations enforced by today's eBPF framework ensure that only expert kernel developers can extend their kernels.

This paper presents \sys, an alternative framework that alleviates the difficulty of writing an eBPF program by allowing Kernel Extensions to be written in Natural language.  \sys uses recent advances in large language models (LLMs) to synthesize an eBPF program given a user's English language prompt.  To ensure that LLM's output is semantically equivalent to the user's prompt, \sys employs a combination of LLM-empowered program comprehension, symbolic execution, and a series of feedback loops.  \sys's key novelty is the combination of these techniques.  In particular, the system uses symbolic execution in a novel structure that allows it to combine the results of program synthesis and program comprehension and build on the recent success that LLMs have shown for each of these tasks individually.

To evaluate \sys, we develop a new corpus of natural language prompts to eBPF programs.  We show that \sys produces correct eBPF programs on \systp---which is an improvement of a factor of \sysimprovetp compared to a LLM-empowered program synthesis baseline.  Moreover, we find that \sys very rarely synthesizes ``false positive'' eBPF programs---i.e., eBPF programs that \sys verifies as correct but manual inspection reveals to be semantically incorrect for the input prompt. The code for KEN is publicly accessible at \href{https://github.com/eunomia-bpf/KEN}{https://github.com/eunomia-bpf/KEN}.
\end{abstract}

\maketitle

\section{Introduction}\label{sec:intro}

Developers are increasingly tasked with modifying and extending operating system kernels to improve performance, security, reliability, or introduce new features to their systems.  Extended Berkeley Packet Filters (eBPF) have emerged as the de facto method for extending an operating system, with recent support for both Linux and Windows~\cite{eBPFWindows}.  eBPF programs inject new logic that is executed before or after existing kernel logic to observe or modify the kernel's behavior.  eBPF programs were originally used to trace network traffic, but the ecosystem now provides sufficient power to implement a variety of features including performance monitoring performance~\cite{gregg16,gregg2021computing}, detecting intrusion detection~\cite{Bachl21,jia2023practical}, and application-specific logic~\cite{Zhong22,yang2023lambda,jia2023programmable,ghigoff2021bmc}.

Unfortunately, eBPF programs are difficult to write correctly.  Implementing an eBPF program requires intimate knowledge of kernel internals to identify where to inject logic~\cite{alibaba23}.  Additionally, the eBPF verifier, intended to prevent unsafe eBPF programs from executing on a system, imposes a number of unfortunate programming constraints on eBPF programmers: programs can only use limited control flow (e.g., loops must have constant bounds) and limited data accesses (e.g., the program cannot access arbitrary memory).  Consequently, eBPF is not even a Turing Complete language~\cite{Mayer21}. 

%\yz{while the verifier prevents several safety issues, it is even harder to [mention semantics issue]...} \arq{If we want to discuss semantic correctness, do it in a new paragraph here.  The issue is that we would want to argue that eBPF is somehow harder to get semantically correct than other programs, but we basically have nothing that really shows this is true} \yz{make sense thx}

In this paper, we present \sys, a system that alleviates the difficulty of eBPF by allowing \underline{\textbf{K}}ernel \underline{\textbf{E}}xtensions written in \underline{\textbf{N}}atural language.  \sys uses recent advances in large language models (LLMs) (e.g.,  OpenAI's InstructGPT\cite{instruct-gpt}, \textit{HumanEval}\cite{human-eval}, and \textit{ChatGPT}\cite{chat-gpt}) to synthesis an eBPF program given a user's English language prompt.  A user can then extend their kernel by injecting the eBPF program produced by \sys without having to understand eBPF or a kernel's internals.

Alas, naively extending a kernel using the output from an LLM is dangerous.  LLMs generate responses that can be unsafe or semantically incorrect.  While the eBPF verifier prevents many unsafe eBPF programs from compromising the kernel, there are fundamental safety issues that remain unpatched~\cite{Jia23}.  Additionally, no existing tools provide assurances that the semantics of \sys's output eBPF program are equivalent to the semantics of the English language prompt input.  Consequently, a naive LLM-based eBPF program synthesis tool only produces a correct program on \basetp~of programs for a representative set of kernel extension prompts.

\sys addresses the semantic correctness challenge by combining three techniques: automated program comprehension, symbolic execution, and feedback-loops.  In addition to using an LLM in its \textit{Synthesis Engine} to synthesize an eBPF program for a user prompt, \sys also uses an LLM in its \textit{Comprehension Engine} to identify Hoare-logic constraint that specifies the conditions that must hold before and after all kernel functions with which an eBPF candidate program interacts.  \sys annotates its synthesized eBPF program with the constraints and passes the results to a symbolic execution engine to validate that the constraints are upheld by the eBPF program.  If a candidate eBPF program passes the symbolic execution engine, \sys further validate its safety using the current eBPF verifier. \sys includes feedback loops to retry synthesis when the system fails to synthesize a verified eBPF program.  It passes error messages back to reprompt itself on failure, which can happen because of a symbolic execution timeout, constraint failure during symbolic execution, or when the eBPF verifier fails.

\sys adopts state-of-the-art techniques for eBPF program synthesis throughout each of its key components (LLM-based program synthesis, LLM-based program comprehension, symbolic execution, and feedback loops).  The system's key novelty lies in its synergistic combination of these techniques. In particular, \sys's use of symbolic execution is a novel structure that allows it to combine the results of program synthesis and program comprehension and build on the success that LLMs have shown for each task individually.

In addition to the system artifact, we produce two novel datasets as a part of this paper.  These datasets are not only useful for building and evaluating \sys, but will also be useful for future work on eBPF program synthesis.  First, \ev is a dataset of eBPF programs and natural language descriptions.  \ev consists of 145 pairs of natural language prompts with eBPF programs.  65 of the paris are from a popular eBPF programming blog~\cite{Gregg01}, while the remaining 80 are handwritten based upon examples from popular open-source eBPF repositories~\cite{bcc,bpftrace,ebpf-exporter,eunomiabpf2023}.  Second, we created \compdataset, a dataset of Hoare-logic contracts for all Linux functions that can be hooked by an eBPF program.  We present a novel methodology for automatically generating such contracts by using the kernel's source code, developer comments embedded in the source code, and an LLM. 

We validate \sys on \ev, using test programs from the blog as a training set and the new tests as the test set.  We find that \sys produces semantically correct eBPF programs on \systp of the test cases, which is \sysimprovetp times better than a naive strategy that only uses LLM-based program synthesis.  Moreover, we find that \sys only produces a single ``false positive''---an output program that passes \sys's verification but is not semantically correct for the provided prompt.  In addition to these high-level results, we show that each of \sys's design principles---feedback-loops, LLM-empowered program comprehension, symbolic execution, and training on the new \ev dataset---contributes to its strong results.  Finally, we explore the performance-cost-privacy tradeoffs between using large cloud-based LLMs (e.g., ChatGPT-4~\cite{chat-gpt}) and small locally-deployable LLMs (e.g., CodeLLama) and show that the large cloud-based LLMs outperform the small LLMs by a factor of 5.3. % This suggests that the improvement in cost and privacy from a small LLM is likely not worth the performance penalty.

%In addition, we show with a case study that \sys's semantic verification approach provides additional safety/security assurances than the existing verifier in eBPF.  In particular, the case study describes an eBPF program that is unsafe (i.e., it can crash the linux kernel), but is nonetheless considered correct by the eBPF verifier.  However, \sys's semantic verification approach identifies the unsafe eBPF program as buggy---the system prevents a kernel crash that would \emph{not} have been caught by the existing eBPF verifier.  This case study suggests that \sys is not only a useful program synthesis tool, but also that its semantic verification could be a useful tool for verifying handcrafted eBPF programs. 

In summary, our contributions are:
\begin{itemize}
    \item We present \sys, the first tool for safely expressing operating system extensions in natural language using program synthesis, program comprehension, symbolic execution, and feedback loops.
    \item We create two datasets: \ev, which provides a rich dataset of eBPF programs and natural language descriptions to aid with training and evaluating program synthesis for eBPF programs, and \compdataset, which provides hoare-logic contracts for Linux helper functions to aid tune program comprehension. 
    \item We evaluate \sys on \ev and show that \sys accurately synthesizes eBPF programs on \systp~of prompts in a representative test set while only verifying a semantically incorrect program in 2.5\% of cases.
\end{itemize}

The rest of this paper proceeds as follows: First, we describe background on each of the techniques employed by \sys (\cref{sec:background}); then, we discuss \sys's design (\cref{sec:design}); next, we describe a case study as an example of how \sys works (\cref{sec:case_study}); we describe our evaluation (\cref{sec:eval}), elaborate on related work (\cref{sec:related}), and concluding (\cref{sec:conclusion}).

%\sys is the first tool employs LLMs to transmute system maintenance instructions rendered in natural language into eBPF programs. To bolster its efficacy, we established a vector database teeming with real-world eBPF programs in \cite{bcc}. Moreover, we amalgamated verification tools and comprehensive kernel function prototypes, accompanied by their detailed descriptions, to curate a "kernel logic guide". Through rigorous testing on real-world maintenance tasks, \sys, when paired with GPT-4, exhibited a commendable average code accuracy of 80\%. This starkly contrasts with the 30\% accuracy achieved using only a handful of examples without our integrated tools. Our curated vector database and automatic retry mechanism further enhanced this figure to 65\%. For perspective, an alternative local-hosted LLM, \textit{code-llama}\cite{roziere2023code}, managed an accuracy of merely 40\%.

\section{Background}\label{sec:background}

This section describes background on the key techniques that \sys employs and a discussion of common eBPF tasks.

\subsection{Key \sys Techniques}
We describe background for each of the key techniques upon which \sys builds. 

\subsubsection{Large Language Models}\label{sec:background:llm} LLMs use machine learning to create sophisticated chat-like programs capable of creating useful responses to natural language prompts. Systems such as \textit{InstructGPT}\cite{instruct-gpt}, \textit{HumanEval}\cite{human-eval}, and \textit{ChatGPT} have employed LLMs for program synthesis, in which a developer prompts the systme with a natural language description of the program and the LLM responds with source code.  Below we describe two standard techniques that have emerged for taking advantage of LLMs.

\paragraph{In-Context Learning}  Systems that employ in-context Learning~\cite{intern-lm,charalambous2023new,liu2023your} adapt an LLM to a new target area by adding minimal contextual data to an LLM (e.g., they prompt the LLM with a small corpus of particularly pertinent inputs) and carefully crafted prompts rather than training a custom LLM on a large dataset.   In-context learning is more cost-effective and can new data more quickly than traditional alternatives.  The key challenges in employing in-context learning in a new synthesis domain involve identifying a corpus of critical examples and crafting a prompt strategy that guides the underlying LLM~\cite{poesia2023certified,charalambous2023new}.

\paragraph{Feedback Loops} Many systems employ a feedback-driven approach to validate an LLM's output against some specification and pass feedback to the model to refine the output.  For example, HarmonyOS developers~\cite{harmony-wiki}, \textsc{Self-Debugging}~\cite{chen2023teaching}, and TPUv4 designers~\cite{tpu} use unit tests to evaluate whether an LLM's output is correct.  The fundamental challenge in developing such ground truth unit tests---manually crafting a set of unit tests that can sufficiently guide the LLM is likely as difficult as manually writing the ideal synthesized output.  Developers in the aforementioned examples were able to use unit tests that were already created for their use cases.  Additionally, \textsc{Self-Debugging} includes a mechanism to provide feedback by using an LLM to explain the behavior of synthesized programs.  The system refines its output by iteratively synthesizing programs, explaining the synthesized program and checking its unit-test outputs, and then iteratively re-synthesizing. 

\subsubsection{Automated Program Comprehension} 
Automated program comprehension determines properties of a program's execution without requiring developer effort.  Early work~\cite{Perkins04} uses a counter-example driven approach, which observe a program under test and derive invariants that hold over all executions.  Recent works shows that LLMs are effective at program comprehension tasks~\cite{Zugner21, Ahmad21, Wu20, feng20, Phan21, Elnaggar21}, but their output is typically an English description of the program.

\subsubsection{Symbolic Execution}
Automated program verification has emerged to ensure that programs are correct without requiring a developer to write a copious amount of tests.  There are a wide range of existing solutions including model checking~\cite{Clarke82, Queille82}, fuzzing~\cite{fuzzing}, and symbolic execution (symbex)~\cite{gurfinkel2015seahorn, klee}.  \sys uses symbex because symbolic execution has been shown to be highly effective in finding issues both in operating systems~\cite{Chipounov11} and user-programs~\cite{klee, gurfinkel2015seahorn}. 

A symbolic execution engine reasons about the behavior of a function or program to determine whether the program upholds specific properties (e.g., memory safety).  A symbex engine associates a symbolic value with each variable in the program.  As the program executes, the engine gathers constraints on the symbolic values (e.g., variable \texttt{x} is less than $3$).  During execution, the engine uses an SMT solver to determine if a given property (e.g., ``can this pointer be equal to NULL?'') can be violated given the constraints that the engine gathered.  Thus, symbolic execution can reason about large classes of inputs to a function or program without needing to execute each input individually.  Symbolic Execution Engines face two fundamental challenges: Path Explosion and Verification Properties. 

\paragraph{Path Explosion} The number of control-flow paths through a program is exponential in the number of conditional operations in the program.  To ensure full coverage of a program, a symbolic execution engines must consider each of these exponentially many paths; this challenge is known as ``the path explosion problem''.  One main consequence of the path explosion problem is that symbolic execution is traditionally limited to analyzing relatively small programs and/or individual functions within programs. 

\paragraph{Verification Properties} Most symbolic execution engines verify that programs satisfy high-level properties such as memory safety (e.g., the program does not dereference a NULL pointer) and safe arithmetic (e.g., the program does not divide by zero).  Some systems have used symbex to ensure that each function satisfies Hoare-logic-style pre- and post-conditions of individual functions~\cite{chipounov2009selective,poeplau2020symbolic}.  

\subsection{eBPF: Extended Berkeley Packet Filter}
\begin{figure}
\fbox{%
\begin{minipage}{\dimexpr\linewidth-2\fboxrule-2\fboxsep}
\textbf{Prompt 1:} If a process tries to use ptrace, it should be killed with signal 2 and its PID should be displayed.\\
\textbf{Prompt 2:} Capture and display all bash commands executed system-wide, along with their exit codes.\\
\textbf{Prompt 3:} If the fan speed is within 90\% of its maximum limit, modify the scheduling policy to optimize performance.
\end{minipage}
}
\caption{Descriptions of typical eBPF management/observability programs.}\label{fig:ebpf-prompts}
\end{figure}

\begin{figure}
    \centering
\begin{minted}[frame=lines,linenos=true]{python}
#!/usr/bin/env bpftrace

tracepoint:syscalls:sys_enter_kill
{
    @tpid[tid] = args.pid;
    @tsig[tid] = args.sig;
}
tracepoint:syscalls:sys_exit_kill
/@tpid[tid]/
{
    time("%H:%M:%S  ");
    printf("%-6d %-16s %-4d %-6d %d\n", pid, comm, @tsig[tid], @tpid[tid],
      args.ret);
    delete(@tpid[tid]);
    delete(@tsig[tid]);
}    
\end{minted}
    \caption{A bpftrace script that implements Prompt 2 from \cref{fig:ebpf-prompts}.}
    \label{fig:bpftrace-example}
\end{figure}

Operating system extension is a critical task for developers today.  Numerous techniques have been developed to ensure safe kernel extensions~\cite{Necula96,Bershad95,Swift03}, with extended Berkeley Packet Filters (eBPF) recently emerging as the de facto standard for today's systems.  eBPF permits custom user-defined program execution within the kernel space without the need to alter the kernel's source or load additional modules. To preserve system security, the eBPF verifier\cite{lwn-verifier} verifies security critical properties of eBPF code prior to its execution.  \Cref{fig:ebpf-prompts} includes three examples of management and observability programs that can be written using eBPF.  In addition to such management tasks, eBPF has been used for scheduling~\cite{humphries2021ghost, pixel6},  security~\cite{Jia23_cfi}, and custom file system rules~\cite{Zhong22}.  For example, Google's Pixel6 kernel~\cite{pixel6} uses eBPF to implement a power-aware real-time scheduler that applies handwritten rules to optimize CPU performance based on user activity.  In one case, the Pixel6 kernel includes an eBPF program that reduces the CPU frequency immediately after a user closes a game. 

The eBPF ecosystem includes multiple frameworks for building tools---the two dominate approaches being \texttt{bpftrace} and \texttt{libbpf}.  \texttt{libbpf} is a C/C++ library that offers a numerous helper functions and structures to use when extending the kernel.  In contrast, \texttt{bpftrace} provides a high-level scripting interface; \cref{fig:bpftrace-example} provides an example bpftrace script for Prompt 2 in \cref{fig:ebpf-prompts} .  can compile \texttt{bpftrace} programs into injectable eBPF bytecode by using the BCC library.  Conventional wisdom is that libbpf implementations are more flexible than bpftrace implementations (i.e., a developer can implement more tasks), but also have more programming complexity~\cite{Gregg20}.   Writing an eBPF program for either framework can be daunting, since a developer needs to understand operating system internals to determine where to inject logic, and is limited in the control flow and data accesses that they can issue.

\section{System Design}\label{sec:design}
\begin{figure}[t]
\centering
\includegraphics[width=0.5\textwidth]{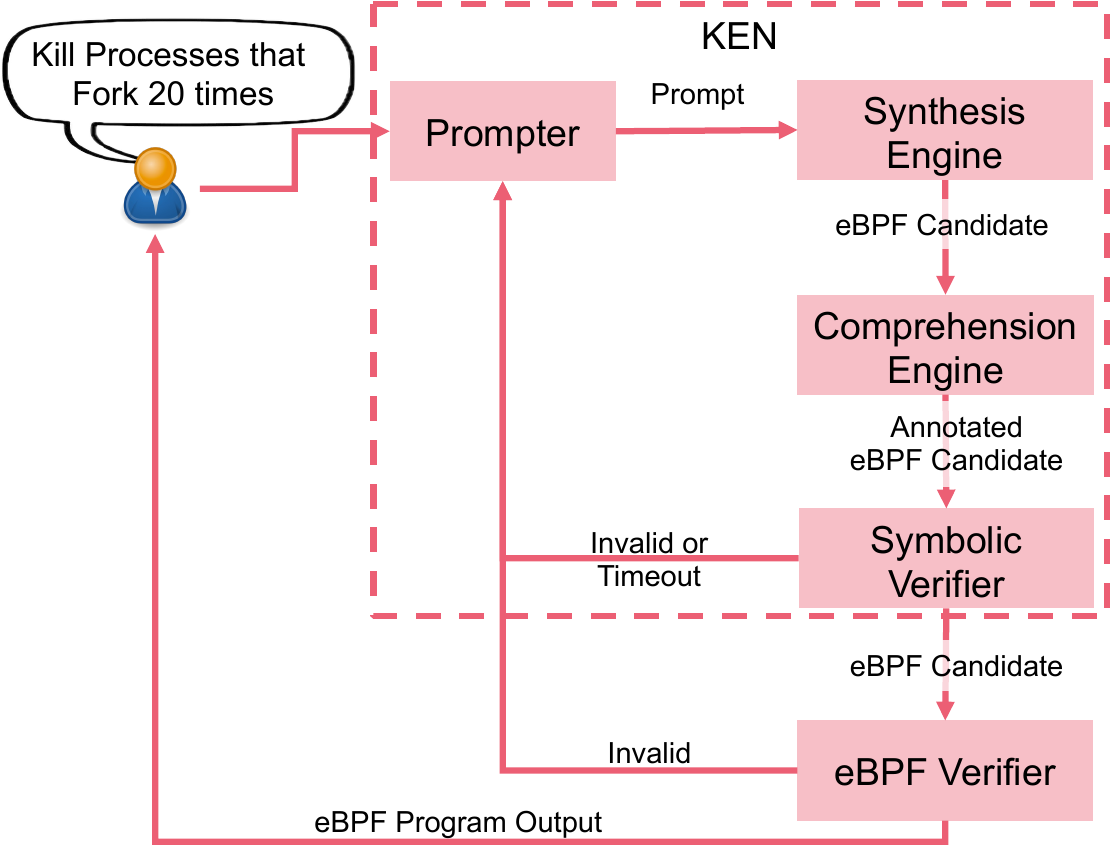}
\caption{The Workflow of \sys}\label{fig:workflow}
\end{figure}

This section describes the design and implementation of \sys; \Cref{fig:workflow} depicts the system's key components and how they interact.  \sys consists of four main components: a Prompter (\cref{sec:design:prompter}), responsible for constructing prompts that generate eBPF programs; a Synthesis Engine (\cref{sec:design:synthesis}), responsible for synthesizing a candidate eBPF program given a natural language prompt; a Comprehension Engine (\cref{sec:design:comprehension}), responsible for annotating a candidate eBPF program with accurate Hoare-logic conditions for each of the kernel functions with which the candidate interacts; and the Symbolic Verifier (\cref{sec:design:symbex}), responsible for ensuring that the candidate eBPF program upholds the Hoare-logic annotations.  The system also uses the existing eBPF verifier to ensure that the eBPF candidate meets basic security criteria.

\sys uses in-context learning (\cref{sec:background:llm}) to augment existing LLMs without requiring training of an entirely new model.  This design allows \sys to remain LLM agnostic---the system can use any LLM and thus supports a variety of cost-performance-privacy tradeoffs (\cref{sec:eval:tradeoffs}).  Supporting in-context learning requires two technical contributions: prompting strategies that guide the LLMs to produce correct output and new datasets for the Synthesis and Comprehension Engines. 

\sys uses a feedback-driven approach to iteratively synthesize a correct eBPF program.  \sys includes two such feedback loops. First, if the Symbolic Verifier determines that a candidate eBPF program from the Synthesis Engine does not uphold the Hoare-logic conditions from the Comprehension Engine or if the Symbolic Verifier times out, then the Verifier will pass the failure back to the Prompter.  This feedback loop allows both Engines additional chances to synthesize correct output.  The second feedback is from the eBPF verifier to the Prompter and is taken when the eBPF verifier does not verify that the synthesized eBPF program is safe. 

Each of \sys's components is an adoption of state-of-the-art techniques to the problem of eBPF program synthesis.  \sys's key novelty lies in its combination of these techniques which empowers a useful and efficient program synthesis tool.  In particular, \sys's use of symbolic execution to combine the results of program synthesis and program comprehension is a novel structure that allows \sys to build on the power that LLMs have shown on both tasks individually. 

The rest of this section proceeds as follows.  We first describe the workflow of synthesizing an eBPF program for a user prompt (\cref{sec:design:workflow}).  Then, we discuss each of \sys's components (\cref{sec:design:prompter}--\cref{sec:design:ebpfverify}).  Finally, we describe \sys's implementation details (\cref{sec:design:implementation}).

\subsection{Workflow}\label{sec:design:workflow}

The high-level workflow for synthesizing an eBPF program in \sys is as follows. The user issues a prompt to the Prompter, which forwards the user's prompt to \sys's Synthesis Engine.  The Synthesis Engine consults an LLM to synthesize a candidate eBPF program based upon the user's input.  \sys passes the candidate eBPF program to the Comprehension Engine, which consults an LLM to annotate the candidate eBPF program with Hoare-logic pre- and post-conditions for each of the Kernel functions that is referenced in the candidate eBPF program.  \sys passes this annotated eBPF candidate to its Symbolic Verifier, which validates that the synthesized eBPF program satisfies the Hoare-logic properties.  If the Symbolic Verifier determines that the eBPF program does not uphold the semantic properties, or if the Symbolic Verifier times out, then the Symbolic Verifier passes its output to the Prompter to begin another iteration of \sys.  If the Symbolic Verifier succeeds, it passes the eBPF program to the eBPF verifier to validate the program's safety properties.  The eBPF verifier will pass its error message to the Prompter to begin another iteration of \sys if the eBPF program is deemed unsafe. 

If \sys fails to synthesize a verified eBPF program after a configured number of trials (3 is the default), the system will re-prompt the user to include additional information.  Anecdotally, we observe that including small additional semantic hints (e.g., hinting at the expected size of some variables) can often resolve \sys's synthesis issues.

\subsection{Prompter}\label{sec:design:prompter}

The Prompter takes the input prompt from the user and specially formats it to pass to the synthesis engine.  In particular, the Prompter adds boilerplate text instructing \sys to produce an eBPF program written for bpftrace framework~\footnote{We experiment with generating eBPF programs written for libbpf in the evaluation (\cref{sec:eval}.}.  Additionally, the Prompter appends all error messages that it has received for the current synthesis task by all feedback loops.  For example, if the Symbolic Verifier failed in a first iteration with message \texttt{failure1} and the eBPF verifier failed in a second iteration with message \texttt{failure2}, then the Prompter will append both \texttt{failure1} and \texttt{failure2} to its message before sending it to the Synthesis Engine.

\subsection{Synthesis Engine}\label{sec:design:synthesis}
%\begin{figure}
%    \centering
%\medskip
%\noindent\fbox{%
%\begin{minipage}{\dimexpr\linewidth-2\fboxrule-2\fboxsep}%
%\textbf{Prompt:}  
%\end{minipage}%
%}
%\caption{\sys Prompt to ensure correct syntax}\label{fig:syntaxControl}
%\end{figure}

The Synthesis Engine takes a natural language prompt and consults an LLM to generate a candidate eBPF program.  The engine uses LangChain~\cite{langchain} as mechanism for interacting with an arbitrary LLM, which allows \sys to support a variety of privacy-cost-performance tradeoffs. 

Like other systems~\cite{chen2023teaching}, the Synthesis Engine uses a VectorDB (e.g., Milvus~\cite{Wang21}) to enable in-context learning.  The engine stores prompt-eBPF pairs from the \ev dataset (see below) in its vectorDB.  On each query, the engine uses the VectorDB to identify the prompt-eBPF pairs that are most similar to the user prompt and includes these pairs as examples of correct input-output pairs in its query to the large language model.  Thus, \sys is similar to few-shot learning~\cite{Wang20}.  \sys also updates its VectorDB after each synthesis, which allows the system to learn from its successful and failure eBPF syntheses.

By specifying the desired syntax in the LLM prompt, we find that the Synthesis Engine almost always synthesizes correct syntax.  Our results show that it also often synthesizes correct semantics (\cref{sec:eval}).  
 
\paragraph{\ev---an eBPF synthesis Dataset}  The Synthesis Engine is empowered by \ev, a novel dataset of 145 natural language prompts paired with corresponding eBPF programs, 79 of which are bpftrace programs and 66 of which are libbpf programs.  \ev was gathered from two sources.  First, 65 pairs (39 bpftrace, 26 libbpf) come from a popular eBPF developer blog~\cite{Gregg01}.  The other 80 pairs (40 bpftrace, 40 libbpf) are hardwritten based upon examples from well-known open-source eBPF project repositories, such as bcc\cite{bcc}, bpftrace\cite{bpftrace}, ebpf-exporter\cite{ebpf-exporter}, or eunomia-bpf\cite{eunomiabpf2023}.  One example pair is Prompt 2 from \cref{fig:ebpf-prompts} and the bpftrace script from \cref{fig:bpftrace-example}. 

\subsection{Comprehension Engine}\label{sec:design:comprehension}

The Comprehension Engines takes an eBPF candidate program as input and annotates it with Hoare-logic pre- and post- conditions at the beginning and end of each of the synthesized eBPF functions.  The engine first uses a regular expression to identify the kernel locations that the candidate program instruments.  The Comprehension Engine obtains the pre- and post-conditions for each kernel location by consulting the \compdataset, which includes pre- and post-conditions that hold for each function that can be instrumented by eBPF (see below).  

The Comprehension Engine prompts an LLM using LangChain.  Its prompt asks the LLM to create assert or assume statements for the beginning and end of each function in the eBPF program, respectively.  The prompt includes the eBPF function, the original developer prompt, and together with the pre- and post-conditions obtained from the \compdataset.  The engine produces output that annotates the candidate eBPF program with LLM-generated assert/assume statements.  Note that \sys could directly use the pre-/post-conditions from the \compdataset to construct assume and assert statements.  However, its output would not be able to use the developer's prompt when generating assume/assert statements and would produce fewer semantically correct programs. Moreover, passing the pre-/post-conditions through the LLM allows \sys to smooth out any inaccuracies in the \compdataset---which arise because the \compdataset is automatically generated and thus neither sound nor complete.  Finally, using an LLM allows the Comprehension Engine to learn from its mistakes and successes, like the Synthesis Engine, by updating an internal VectorDB on every prompt.

It is worth noting that the LLM in a Comprehension Engine could nefariously adjust its pre-/post-conditions so that all candidate eBPF programs are tautologically verified by the symbolic verifier.  Our evaluation (\cref{sec:eval}) shows empirically that this is not the case since \sys synthesizes few ``false positives''---i.e., programs that the system verifies but that manual inspect reveals to not implement the same semantics as the natural language prompt.  We hypothesize that one reason that we do not see the Comprehension Engine gaming the system towards tautology is that the engine's VectorDB learns slowly and so we do not observe the VectorDB drastically impacting the LLM behavior.  It is possible that a extremely long-running deployment would observe more such false positives. This issue could then be resolved by occasionally clearing the Comprehensive Engine's VectorDB.

\paragraph{\compdataset---a dataset of Hoare-logic contracts for kernel functions}
Manually creating a dataset of Hoare-logic contracts for every eBPF instrumentable kernel function would be infeasible.  There are hundreds of such functions, and the functions themselves change with each new kernel release.  Thus, we chose to generate \compdataset using an automated approach that we can then adapt to new kernel versions.  We use a regular expression to identify every function that has a symbol in the kernel.  To approximate the function's semantics, we use a regular expression to find any comments immediately before the function.  We pass the function prototype, source code, and approximate semantics into an LLM using a prompt that asks for Z3 compatible conditions for each of the eBPF helpers.  We store the approximate semantics, prototype, and the output from the LLM in a JSON format in the \compdataset.  If detected, developers could fix inaccuracies from the automated approach, although we have not needed to do so in \sys.

\subsection{Symbolic Verifier}\label{sec:design:symbex}

The symbolic verifier uses symbolic execution to validate the annotated eBPF candidate program produced the Comprehension Engine.  If the symbolic verifier determines that the program upholds the assert statements, it removes the assert/assume statements and passes the candidate eBPF program to the eBPF verifier.  If the symbolic verifier finds an assertion statement that is not upheld or times out, then it passes its error message to the Prompter. 

\subsection{eBPF Verifier}\label{sec:design:ebpfverify}

\sys uses the existing eBPF Verifier from the operating system.  Namely, the eBPF Verifier validates that the eBPF candidate program produced by the Symbolic Verifier contains no unbounded loops or arbitrary memory accesses.  If the eBPF verifier is unable to assure that the candidate program is safe, it passes its error message back to the Prompter.  Otherwise, the eBPF verifier passes the eBPF candidate program to the user as its final synthesized eBPF program.

\subsection{Implementation}\label{sec:design:implementation}

We implement \sys, \ev, and \compdataset using 4244 LOC in Python and 490 LOC cumulatively acrosss HTML, CSS, and JS.  In addition, we add 51 LOC to the bpftrace compiler to add support for assume and assert functions.  \sys uses SeaHorn \cite{gurfinkel2015seahorn} as its Symbolic Verifier and Z3~\cite{DeMoura08} as its backend SMT solver.  SeaHorn symbolically executes LLMV IR; \sys uses SeaHorn by compiling the output of the Comprehension Engine to LLVM IR.

\sys supports a multitude of LLM, but employs ChatGPT-4 by default.  While eBPF is now supported in Windows, our current \sys prototype only works on Linux.  \sys can synthesize eBPF programs for both libbpf and bpftrace, but it defaults to bpftrace because we found that \sys is more effective at synthesizing bpftrace programs (\cref{sec:eval}).

\section{Case Study}\label{sec:case_study}

This section describes a case study of using \sys to provide an example of how each of \sys's steps work.  We crafted this scenario specifically to illustrate \sys's behavior; it is not a test from \ev.

In this case study, a user wishes to display information to their terminal about all of the tcp connections that are made to their webserver so that they can detect denial of service attacks that are made against it. Kernel extensions offers a good mechanism for performing full system telemetry.  However, since the developer is not a linux kernel expert, they turn to \sys to help them build their eBPF program.  \Cref{fig:case:prompt} shows the developer's prompt that the developer issues to \sys.  They ask the system to trace all IPv4 and IPv6 tcp\_connect events and specify their desired output including the proper order for each port.

\begin{figure}[t]
\begin{framed}
\raggedright
\textbf{Prompt:} Write a bpftrace program to \sethlcolor{bggray}\hl{trace tcp\_connect events}\sethlcolor{yellow}\hl{ for both IPv4 and IPv6 connection attempts}, \sethlcolor{bggray}\hl{display the source and destination IP addresses} \sethlcolor{bluegray}\hl{ and the source and destination ports in host byte order}.
\end{framed}
\caption{A prompt passed into \sys instructing it to print basic connection information for all tcp connect attempts.}\label{fig:case:prompt}
\end{figure}

\begin{figure}[t]
\begin{framed}
\begin{minted}[fontsize=\tiny,linenos=false,escapeinside=@@,highlightlines={1-18},autogobble=true,breaklines=true]{c++}
@\xglobal\colorlet{FancyVerbHighlightColor}{bggray}@kprobe:tcp_connect {
@\xglobal\colorlet{FancyVerbHighlightColor}{bggray}@    $saddr = ntop(2, $sk->__sk_common.skc_rcv_saddr);
    $daddr = ntop(2, $sk->__sk_common.skc_daddr);
    $sport = ($sk->__sk_common.skc_num);
    $dport = ($sk->__sk_common.skc_dport);
    printf("TCP connect: %s:%d -> %s:%d\n", $saddr, $sport, $daddr, $dport);
@\xglobal\colorlet{FancyVerbHighlightColor}{bggray}@}
\end{minted}
\end{framed}
\caption{The output of the Synthesis Engine on the prompt from \cref{fig:case:prompt}.}\label{fig:case:synthesis}
\end{figure}

\begin{figure}[t]
\begin{framed}
\begin{minted}[fontsize=\tiny,linenos=false,highlightlines={3},highlightcolor=bluegray]{javascript}
"kretprobe:tcp_connect_init": {
    "pre": {
        "sk": "!=null",
    }
}
\end{minted}
\end{framed}

\caption{JSOn for the matching element in \compdataset for the kernel probe specified in \cref{fig:case:synthesis}.}\label{fig:case:spec}
\end{figure}

\begin{figure}[t]
\begin{framed}
\begin{minted}[fontsize=\tiny,linenos=true,escapeinside=@@,highlightlines={1-18},autogobble=true,breaklines=true]{c++}
@\xglobal\colorlet{FancyVerbHighlightColor}{bggray}@kprobe:tcp_connect {
    $sk = (struct sock *) arg0;
@\xglobal\colorlet{FancyVerbHighlightColor}{brightpink}@    assume($sk != 0);
    assume($sk->__sk_common.skc_rcv_saddr != 0);
    assume($sk->__sk_common.skc_daddr != 0);
    assume($sk->__sk_common.skc_num >= 0);
    assume($sk->__sk_common.skc_dport >= 0);
@\xglobal\colorlet{FancyVerbHighlightColor}{yellow}@    assume(sizeof($sk->__sk_common.skc_rcv_saddr) == 4 || sizeof($sk->__sk_common.skc_rcv_saddr) == 16);
    assume(sizeof($sk->__sk_common.skc_daddr) == 4 || sizeof($sk->__sk_common.skc_daddr) == 16);
@\xglobal\colorlet{FancyVerbHighlightColor}{bggray}@    $saddr = ntop(2, $sk->__sk_common.skc_rcv_saddr);
    $daddr = ntop(2, $sk->__sk_common.skc_daddr);
    $sport = ($sk->__sk_common.skc_num);
    $dport = ($sk->__sk_common.skc_dport);
    printf("TCP connect: %s:%d -> %s:%d\n", $saddr, $sport, $daddr, $dport);
@\xglobal\colorlet{FancyVerbHighlightColor}{bluegray}@    assert($dport == bswap($sk->__sk_common.skc_dport));
@\xglobal\colorlet{FancyVerbHighlightColor}{bluegray}@    assert($sport == bswap($sk->__sk_common.skc_num));
@\xglobal\colorlet{FancyVerbHighlightColor}{bggray}@}
\end{minted}
\end{framed}
\caption{Annotated candidate eBPF program produced by the Comprehension Engine for verification. The original synthesized eBPF program is shown in gray. Pre-conditions inferred from \compdataset are shown in pink. Pre-conditions inferred from the user prompt are shown in yellow. Finally, post-conditions inferred from the user's prompt are shown in blue.}\label{fig:case:comprehension}
\end{figure}

\begin{figure}[t]
\begin{framed}
\begin{minted}[fontsize=\tiny,linenos=false,escapeinside=@@,highlightlines={1-18},autogobble=true,breaklines=true]{c++}
@\xglobal\colorlet{FancyVerbHighlightColor}{bggray}@kprobe:tcp_connect {
@\xglobal\colorlet{FancyVerbHighlightColor}{bggray}@    $saddr = ntop(2, $sk->__sk_common.skc_rcv_saddr);
    $daddr = ntop(2, $sk->__sk_common.skc_daddr);
    $sport = (bswap($sk->__sk_common.skc_num));
    $dport = (bswap($sk->__sk_common.skc_dport));
    printf("TCP connect: %s:%d -> %s:%d\n", $saddr, $sport, $daddr, $dport);
@\xglobal\colorlet{FancyVerbHighlightColor}{bggray}@}
\end{minted}
\end{framed}
\caption{The output of the Synthesis Engine on the second iteration for the prompt of \cref{fig:case:prompt} and the error message for the symbolic verification failure of \cref{fig:case:comprehension}}\label{fig:case:synthesis2}
\end{figure}

\sys Prompter passes the user's prompt to the Synthesis Engine, which generates the output program shown in \cref{fig:case:synthesis}.  The program parses the relevant data from the \texttt{sk} parameter of the kernel probe \texttt{tcp\_connect}.  The Synthesis Engine passes the synthesized candidate eBPF program to the Comprehension Engine.  

The Comprehension Engine first uses a regular expression to lookup any pre- and post-conditions in its \compdataset for the \texttt{tcp\_connect} kprobe. \Cref{fig:case:spec} shows the matching pre-condition.  The Comprehension Engine then prompts its LLM to identify assert and assume statements for the candidate eBPF program, passing input as the matching pre-condition, the original user prompt, and the candidate eBPF program.  \Cref{fig:case:comprehension} shows the annotated eBPF candidate program generated by adding the LLM generated assume and assert statements.  The LLM generates the assume in pink based upon the pre-condition from the \compdataset element.  The LLM generates the yellow assume statements based upon the user prompt.  Finally, the LLM generates the assert statements in blue based upon the users prompt.  Note that the final statement determines that bytes should be swapped because the user's query asks for host byte order. 

Then, \sys passes the annotated eBPF candidate program to the symbolic verification engine.  The engine translates the assume an assert statements into SMT formulas and passes them to Z3 to verify if the originally generated eBPF program satisfies both the safety requirements from the eBPF verifier and semantics requirements from the prompt.  Specifically, since line 13 contradicts line 15 in \cref{fig:case:comprehension}, and line 12 contradicts line 16, the symbolic verification engine identifies the program as invalid.   

\sys's feedback loops perform another iteration of synthesis, adding error messages indicating the failed assert statements to the prompt.  This time, the Synthesis Engine creates the synthesized program in \cref{fig:case:synthesis2}, with \texttt{sport} and \texttt{dport} variables correctly assigned in host byte order.  The Comprehension Engine identifies the same assume and assert statements as were identified in \cref{fig:case:comprehension}.  On this iteration, the new candidate program is passed to the symbolic verifier, the verifier determines that the candidate eBPF program is correct.  Furthermore, the eBPF verifier confirms that the candidate contains no security issues, so the developer can safely extend their kernel with the new program.

\section{Evaluation}\label{sec:eval}

We implemented a prototype of \sys and used it to answer the following research questions: 

\begin{itemize}
\item \textit{RQ1} How effective is \sys at synthesizing eBPF programs for representative eBPF prompts?
\item \textit{RQ2} How does each aspect of \sys's design contribute to the system's accuracy?
\item \textit{RQ3} What are the performance, privacy, and cost trade-offs from using different LLM implementations for \sys? 
\end{itemize}

\sys employs ChatGPT4 as its LLM unless otherwise specified.  Additionally, unless otherwise specified, we train the system using \compdataset and the set of prompt-eBPF program pairs from \ev that were scrapped from the popular web blog~\cite{Gregg01} and test the system using the set of prompt-eBPF program pairs from \ev that we created anew.  We create a baseline that synthesizes eBPF programs using a single prompting of ChatGPT-4 and verifies the output eBPF program by using the built-in eBPF verifier. 

Since each prompt can be correctly implemented in multiple ways, we manually inspect \sys's synthesized eBPF program for each prompt to determine if the output correctly implements the prompt.  We calculate the \emph{Accuracy (A)} of \sys for each experiment, which is the fraction of prompts for which \sys synthesized a correct eBPF program. 

To understand the consequence of \sys's incorrect outputs, we split the prompts for which \sys fails to correctly synthesize an eBPF program into two categories: \emph{False Negative (FNs)}, which are the percentage of prompts for which \sys fails to synthesize a verified eBPF program, and \emph{False Positives (FPs)}, which are the percentage of prompts for which \sys synthesizes a verified eBPF program that does not correctly implement the prompt.  Conceptually, FPs represent a safety violation since a developer using \sys may extend their kernel incorrectly when \sys produces a false positive.  In contrast, FNs represent a liveness violation since a developer is effectively unable to use \sys for such prompts.

\subsection{RQ1: \sys Effectiveness}
\begin{figure}[t]
\includegraphics[width=0.5\textwidth]{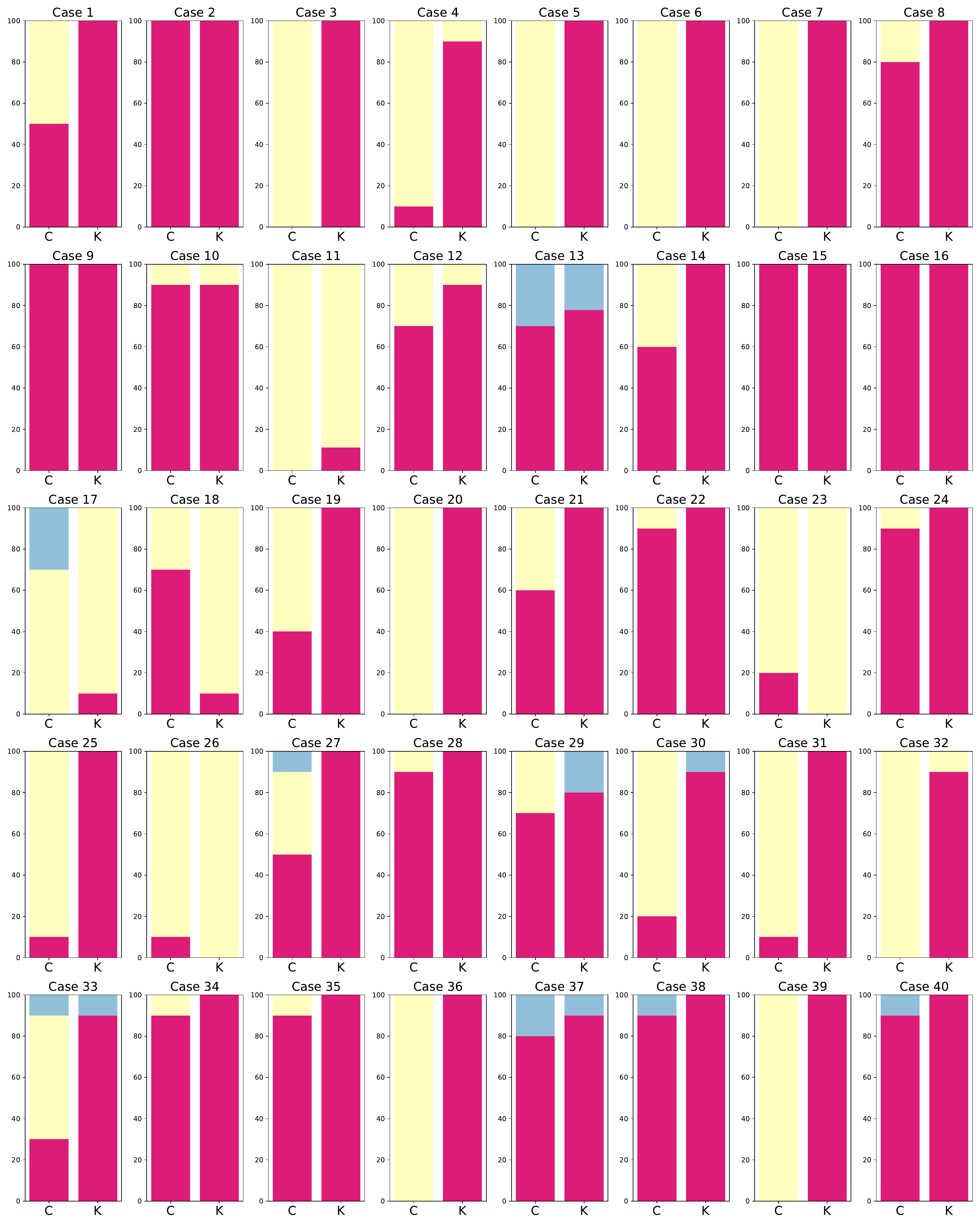}
\caption{The per-prompt effectiveness of \sys (K) and a ChatGPT-4 (C).  Each bar chart shows the percentage of time that \sys/ChatGPT-4 synthesizes an accurate(pink), false negative (yellow), or false positive(blue) eBPF program for each prompt over ten trials.}\label{fig:histogram}
\end{figure}

\begin{table}[t]
 \caption{End-to-End Effectiveness comparison of \sys with a ChatGPT-4 baseline} \label{tab:endToEnd}
    \centering
    \begin{tabular}{|l|c|c|c|}  \hline
        \textbf{System}  & A & FP & FN \\ \hline
        \textbf{ChatGPT-4} & 30\% & 2.5\% & 67.5\%  \\
        \textbf{\sys} & 80\%  & 2.5\% & 17.5\% \\\hline
    \end{tabular}
\end{table}

For each prompt in the \ev testing set,  \cref{fig:histogram} shows a bar depicting the percentage of time that \sys and the ChatGPT-4 baseline are accurate, produce a false positive, and produce false negative across 10 iterations of the prompt.  \Cref{tab:endToEnd} shows the average number of prompts on which each system is accurate, has a false positive, and produces a false negative for each trial.   On average, \sys achieves an accuracy of \systp, which is a factor of \sysimprovetp better than the ChatGPT-4 accuracy of \basetp.  Moreover, \sys achieves the accuracy improvement without increasing the False Positive rate---both systems achieve an FP rate of only 2.5\%.

Inspecting the results for each individual test case, we observe that \sys generates a correct program more often than the baseline in 37 of the 40 test cases.  In 11 of 40 test cases, \sys improves the accuracy rate by more than a factor of 9 (i.e., the accuracy rate improves from at most 10\% in the baseline to at least 90\% in \sys).  In addition, \sys improves on the false positive rate in 6 of the 9 test cases in which either system observes a false positive. 

\subsection{RQ2: The Effectiveness of \sys's Design Decisions}

We first evaluate the impact of each of \sys's high-level design decisions, then describe an experiment showing the benefit of \sys's comprehension and symbolic execution engines, and finally describe the impact of synthesizing eBPF programs to use bpftrace instead of libbpf. 

\subsubsection{Effectiveness of \sys's High-Level Design Decisions}
\begin{table}[t]
    \caption{Comparison of effectiveness for each of \sys's design decisions}
    \centering
    \begin{tabular}{|l|c|c|c|}  \hline
        \textbf{System}  & A & FP & FN \\ \hline
        \textbf{ChatGPT-4} & 30\% & 2.5\% & 67.5\%  \\
        \textbf{ChatGPT-4+Feedback} & 60\% & 7.5\% & 32.5\%  \\
        \textbf{ChatGPT-4+Feedback+Symbex} & 77.5\% & 5\% & 17.5\%  \\
        \textbf{\sys} & 80\%  & 2.5\% & 17.5\% \\\hline
    \end{tabular}
    \caption{The Breakdown Accuracy Analysis of \sys}\label{tab:breakdown}
\end{table}
Table~\cref{tab:breakdown} shows how each high-level design decision impacts \sys's effectiveness.  Each row in the table represents a different configuration. We start from the ChatGPT-4 baseline and apply \sys's design features in the the order of largest impact on accuracy.  Namely, we first include \sys's model-guided feedback (feedback), then \sys's comprehension and symbolic execution components (symbex), and finally add training using the blog-gathered portion of the \ev dataset.  We note that it is not meaningful to separate \sys's comprehension engine from its symbolic execution components.

The results indicate that model-guided feedback plays a large role in improving the accuracy of \sys, as it improves the accuracy by a factor of 2 (from \basetp to 60\%).  However, this increase in accuracy comes with a factor of 3 increase in false positive rate (from 2.5\% to 7.5\%).  Including the comprehension engine and symbolic execution component also improves \sys's effectiveness substantially---accuracy improves to 77.5\%, while the false positive rate moves to 5\%.  Including the \ev dataset in training comes with a relatively small impact on \sys's accuracy---it only improves by 2.5\%.  However, training using the \ev dataset does bring \sys's false positive rate back down to the baseline of 2.5\%

\subsubsection{Effectiveness of \sys's Comprehension and Symbolic Execution Engine}

We create a baseline that replaces \sys's automated reasoning components (i.e., the comprehension and symbolic execution engines) with developer expertise to identify the power of \sys's automated reasoning.  In particular, we augment the \ev testset by adding the correct kprobe and kretprobe locations for each of the prompts in the \ev testset.  Then, we produce a \emph{human expertise} baseline by using the augmented dataset as input to the ChatGPT-4 + feedback + dataset baseline.  

\cref{tab:humanExpertise} shows the effectiveness of the human expertise baseline compared to \sys.  We find that \sys has higher accuracy, without adding additional false positives, compared to the human expertise baseline.  \sys's superiority is particularly surprising since \sys does not directly pass the output of the comprehension and symbolic execution engines into its synthesis engine.  Our hypothesis is that \sys's feedback loop, initiated on each symbolic execution failure, provides the synthesis engine with sufficient knowledge to replace the human expertise from the baseline.  Thus, we find a powerful synergy between \sys's feedback feature and its automated reasoning features. 

\begin{table}[t]
 \caption{End-to-End Effectiveness comparison of \sys with a ChatGPT-4 baseline}\label{tab:humanExpertise}
    \centering
    \begin{tabular}{|l|c|c|c|}  \hline
        \textbf{System}  & A & FP & FN \\ \hline
        \textbf{Human Expertise} & 72.5\% & 2.5\% & 25\%  \\
        \textbf{\sys} & 80\%  & 2.5\% & 17.5\% \\\hline
    \end{tabular}
       
\end{table}

\subsubsection{Effectiveness of using bpftrace instead of libbpf}
We also evaluate the impact of synthesizing programs that use bpftrace compared to synthesizing programs using libbpf.  Conventional wisdom is that libbpf implementations are more flexible than bpftrace implementations (i.e., a developer can implement more tasks), but also have more programming complexity~\cite{Gregg20}.  We evaluate \sys's ability to synthesize eBPF programs that use bpftrace and libbpf.  However, we find that \sys's symbolic execution engine rarely terminates on libbpf programs because of state explosion due to the frequently use of helper function in libbpf programs.  Nonetheless, we calculate the accuracy of \sys when we do not use \sys's comprehension and symbolic execution engines and determine that the accuracy of libbpf programs is 37.5\%, whereas the accuracy of bpftrace programs is 60\%.  Additionally, we calculate the latency of each \sys prompt and find that libbpf programs take an average of 2.16 seconds to synthesize, whereas \sys can synthesize bpftrace programs in an average of 1 second.  We conclude that bpftrace is a better synthesis target for our eBPF use cases.

\subsection{RQ3: The Performance-Cost-Privacy tradeoffs of alternative LLM models}\label{sec:eval:tradeoffs}

In this section, the performance of \sys is assessed across several recent Large Language Models (LLMs), each presenting distinct potentials for privacy leakage and execution time cost tradeoffs since we expect running locally won't hurt the experience and lower the accuracy that much. We aim to deconstruct both CPU and GPU time to determine the extent of time savings achieved by offloading to remote, given that the CPU time is predominantly consumed by Z3 and the GPU time is primarily allocated for inference. The system is executed on a commercially available Intel 12900K with 96GB DRAM installed and Nvidia Geforce 4090 with 24GB VRAM. We are utilizing the Python profiling tools \texttt{SCALENE} \cite{berger2023triangulating} for getting the Max VRAM occupation and CPU GPU time for different models, CodeLLama\cite{roziere2023code},  WizardLM\cite{luo2023wizardcoder} with 7B and 13B parameter size in Fig.~\ref{tab:localdp}.

\begin{table}[h]
     \caption{Comparison with different locally deployed LLM}\label{tab:localdp}
    \centering
    % \hspace{-0.5em}
\begin{tabular}{|>{\centering\arraybackslash}l|m{3em}|m{5em}|m{5em}|}
    \hline
    \textbf{Type} & \textbf{TP \%} &\textbf{Inference time}&\textbf{Max VRAM(GB)} \\ 
    \hline
    \textbf{CodeLLama-7B} & 12.5\% &  3m22.971s &7.32 \\
    \textbf{CodeLLama-13B} & 15\% &  5m55.860s &8.90 \\
    \textbf{WizardLM-7B} & 5\% & 10m3.396s & 18.03\\
    \textbf{WizardLM-13B} & 12.5\%  & 46m37.193s & 17.96\\
    \hline
\end{tabular}
\end{table}

Since the Z3 CPU time is interleaved within other thread waiting times in the Python profiler, we estimate the time function-wise by running the overall Z3 function calling time and Inference function time for one query. We found that Z3 running time corresponds to the \texttt{assume} and \texttt{sassert} number and takes less than 10s for most of the queries. While the 13B model can be accommodated within the 4090 VRAM, the 7B model proves to be more practical owing to its superior inference speed.  The 4090 is deployed with fp8 inference quantization using transformers pipeline as a high-level helper. For CodeLLama, we use model "TheBloke/CodeLlama-Instruct-GPTQ"; for WizardLM, we use "WizardLM/WizardCoder-Python-V1.0". We observed that when WizardLM models are deployed locally, VRAM is almost completely utilized. The transformer pipeline alternates between memory and VRAM to accommodate the model size for inference. We notice a performance gain of 1.2× TP in CodeLLama but at the cost of 1.5× more time, and in WizardLM, a gain of 2.5× TP occurs but with a 4.6× increase in time cost. The suboptimal performance observed with fewer parameters is due to the model easily losing context and generating incorrectly formatted text. Compared with CodeLLama-34B's 42.5\%, which is 3x-6x of the locally deployed ones, we think offloading LLM to the remote servers is a better solution.

\section{Related Work}\label{sec:related}
\sys builds on related work in synthesis, verification, and large language models.   This section focuses its discussion on related work that has not been discussed elsewhere in the paper; \cref{sec:background} provides more context for the individual techniques that \sys employs.

\sys builds on decades of work into program synthesis~\cite{SolarLezama06, Jha10, Torlak13, Alur18,Gulwani11}, using large language models as advocated by many recent works~\cite{austin21, feng20, chen21, Li22}.  \sys's key differences from this prior work is its novel use of a program comprehension and symbolic execution to add assurance that synthesized programs are semantically correct.  One key reason for this differences is that \sys targets kernel extension, in which an incorrectly synthesized program can have extremely bad consequences.  One additional difference between our work and these prior works is that eBPF is significantly less popular compared to the languages tested by existing work (e.g., python), so our work necessitated developing new datasets. 

Recently, \textsc{Self-Debugging} integrated Large Language Models with feedback loops to synthesize programs~\cite{chen2023teaching}.  The system has a synthesis and feedback structure to \sys.  However, the system uses unit tests and code explanation modules to validate synthesized programs instead of \sys's symbolic execution. Additionally, \textsc{Self-Debugging} targets python and SQL synthesis instead of \sys's eBPF target. 

Automated program comprehension determines properties of a program's execution without requiring developer effort.  Early work~\cite{Perkins04} uses a counter-example driven approach, which observe a program under test and derive invariants that hold over all executions.  Such systems require a large corpus of test to ensure coverage of the invariants in the program;  \sys's approach instead learns less accurate invariants from the original source code. Recent works shows that LLMs are effective at program comprehension tasks~\cite{Zugner21, Ahmad21, Wu20, feng20, Phan21, Elnaggar21}.  However, the output of existing LLM-based tools is typically an English description of the program which cannot be passed to program verification tools such as symbolic execution. 

Like \sys, many systems have embraced machine learning and large language models to enhance developer tools.  For example, Neural Fuzzing~\cite{sun2021healer} uses a neural network trained to identify bugs in operating systems more effectively and improves runtime coverage compared to traditional coverage-guided fuzzing tools.  As another example, Scalene~\cite{berger2023triangulating} recently began producing LLM-synthesized patches for performance bugs that are found by the profiler. 

Many prior works have identified and tried to address verification issues in eBPF.  Jia et al. identify challenges that arise in existing eBPF verification due to the large use of helper functions in modern systems~\cite{Jia23}.  The authors advocate for the use of safe languages (e.g., Rust) to build future kernel extensions.  \sys takes an alternative approach to improve on safety by using symbolic execution and Hoare-logic contracts.  K2~\cite{Xu21} uses synthesis techniques to optimize the performance of eBPF programs that are passed to the system while ensuring that the optimized programs still pass eBPF's verifier.  Thus, whereas \sys's semantic verification provides additional assurances over the eBPF verifier, K2 provides the same safety assurances as the eBPF verifier.

\section{Conclusion}\label{sec:conclusion}

In conclusion, we presented \sys, a system that helps developers extend their kernels using eBPF by allowing \underline{\textbf{K}}ernel \underline{\textbf{E}}xtensions written in \underline{\textbf{N}}atural language.  \sys uses recent advances in large language models (LLMs) (e.g.,  OpenAI's InstructGPT\cite{instruct-gpt}, \textit{HumanEval}\cite{human-eval}, and \textit{ChatGPT}\cite{chat-gpt}) to synthesis an eBPF program given a user's English language prompt.  A user can then extend their kernel by injecting the eBPF program produced by \sys without having to understand eBPF or a kernel's internals.  We show that \sys can addresses the challenge of generating a semantically correct eBPF program for a given prompt by combining three techniques: LLM-based program comprehension, symbolic execution, and feedback-loops.  In particular, \sys uses an LLM in its \textit{Comprehension Engine} to identify Hoare-logic constraint that specifies the conditions that must hold before and after all kernel functions with which an eBPF candidate program interacts.  It then uses a symbolic execution engine to validate that the constraints are upheld by the eBPF program and the eBPF verifier to further ensure the safety of the candidate program.  In the event that a program cannot be verified, the system retires the synthesis task by passing error message to an additional round of synthesis.

\sys's key novelty lies in its synergistic combination of state-of-the-art techniques---LLM-based program synthesis, LLM-baesd program comprehension, symbolic execution, and feedback loops.  In particular, \sys innovates with a new architecture that uses symbolic execution to combine the results of program synthesis and program comprehension.  Thus, the system creates a synthesis approach that build on the success that LLMs have shown for each task individually.

In addition to \sys, we produce two novel datasets that will be not only useful for building and evaluating \sys, but will also be useful for future work on eBPF program synthesis.  First, \ev is a dataset of eBPF programs and natural language descriptions gathered from a popular blog and popular open-source repositories.  Second, we created \compdataset, a dataset of Hoare-logic contracts for all linux functions that can be hooked by an eBPF program.  We presented a novel methodology for automatically generating such contracts by using the kernel's source code, developer comments embedded in the source code, and an LLM. 

Lastly, we validated \sys on \ev.  We conclude the following from our evaluation: 

\begin{enumerate}
    \item \sys's novel architecture produces semantically correct eBPF programs significnalty more often than LLM-based program synthesis alone.  In particular, \sys synthesizes correct eBPF programs in \systp of the test cases, which is \sysimprovetp times better than an LLM-based program synthesis baseline.
    \item \sys's novel architecture prevents the tool from being overly confident.  In particular, \sys only produces a single ``false positive''---an output program that passes \sys's verification but is not semantically correct for the provided prompt.
    \item We show that each of the aspects of \sys's novel architecture contribute to its strong results. 
    \item We show that the reduced cost and improved privacy of using a small locally-deployable LLM compared to a large cloud-based LLM is likely not worth the performance cost.  In particular, we show that \sys deployed on ChatGPT-4 produces accurate results more frequently than \sys deployed on CodeLlama-13B by a factor of 5.3.
\end{enumerate}

\begin{comment}
    
\paragraph{The code example guide the vectorDB moves the accuracy to 65\%}
Our results highlight that the presence of similar code examples in vectorDB can increase accuracy rates to around 80\%. Without this foundational knowledge in vectorDB, it remains a challenge to expand its understanding to include all potential code snippet combinations.

\paragraph{The ground truth guided way of re-prompt to 80\% }
While LLMs can sometimes produce misleading or "hallucinated" outputs, guiding the model with a ground truth reference can help enhance its accuracy. Such guidance addresses potential pitfalls and ensures the model achieves to 80\% accuracy, addressing the remaining 15\% gap.

\paragraph{Comparative Assessment of libbpf and bpftrace}
The head-to-head comparison of libbpf and bpftrace reveals distinct performance characteristics for each tool. While bpftrace outperforms libbpf in certain benchmarks, the exact trade-offs in accuracy and efficiency between the two tools are contingent upon specific use-cases and datasets.

\paragraph{Deployment Feasibility on Local machine}
Our evaluations suggest that z3 can be run on local machines, while deploying the LLM model on a local 4090 is insufficient for direct use, it requires at least a 34B parameter size to memorize long enough context for in-context learning.
\end{comment}
\section{Acknowledgement}\label{sec:acknowledgement}
We acknowledge the help of formal description for the guided example from Zhe Ye.

\bibliographystyle{abbrvnat}

\bibliography{cite}

\begin{thebibliography}{67}
\providecommand{\natexlab}[1]{#1}
\providecommand{\url}[1]{\texttt{#1}}
\expandafter\ifx\csname urlstyle\endcsname\relax
  \providecommand{\doi}[1]{doi: #1}\else
  \providecommand{\doi}{doi: \begingroup \urlstyle{rm}\Url}\fi

\bibitem[Ahmad et~al.(2021)Ahmad, Chakraborty, Ray, and Chang]{Ahmad21}
W.~Ahmad, S.~Chakraborty, B.~Ray, and K.-W. Chang.
\newblock Unified pre-training for program understanding and generation.
\newblock In \emph{Proceedings of the 2021 Conference of the North American
  Chapter of the Association for Computational Linguistics: Human Language
  Technologies}, pages 2655--2668, Online, June 2021. Association for
  Computational Linguistics.
\newblock \doi{10.18653/v1/2021.naacl-main.211}.
\newblock URL \url{https://aclanthology.org/2021.naacl-main.211}.

\bibitem[Alur et~al.(2018)Alur, Singh, Fisman, and Solar-Lezama]{Alur18}
R.~Alur, R.~Singh, D.~Fisman, and A.~Solar-Lezama.
\newblock Search-based program synthesis.
\newblock \emph{Commun. ACM}, 61\penalty0 (12):\penalty0 84–93, nov 2018.
\newblock ISSN 0001-0782.
\newblock \doi{10.1145/3208071}.
\newblock URL \url{https://doi.org/10.1145/3208071}.

\bibitem[Austin et~al.(2021)Austin, Odena, Nye, Bosma, Michalewski, Dohan,
  Jiang, Cai, Terry, Le, and Sutton]{austin21}
J.~Austin, A.~Odena, M.~Nye, M.~Bosma, H.~Michalewski, D.~Dohan, E.~Jiang,
  C.~Cai, M.~Terry, Q.~Le, and C.~Sutton.
\newblock Program synthesis with large language models, 2021.

\bibitem[Authors(2023)]{eunomiabpf2023}
E.~B. Authors.
\newblock Eunomia bpf.
\newblock GitHub repository, 2023.
\newblock \url{https://github.com/eunomia-bpf/eunomia-bpf}.

\bibitem[Bachl et~al.(2021)Bachl, Fabini, and Zseby]{Bachl21}
M.~Bachl, J.~Fabini, and T.~Zseby.
\newblock A flow-based {IDS} using machine learning in ebpf.
\newblock \emph{CoRR}, abs/2102.09980, 2021.
\newblock URL \url{https://arxiv.org/abs/2102.09980}.

\bibitem[Berger et~al.(2023)Berger, Stern, and
  Pizzorno]{berger2023triangulating}
E.~D. Berger, S.~Stern, and J.~A. Pizzorno.
\newblock Triangulating python performance issues with $\{$SCALENE$\}$.
\newblock In \emph{17th USENIX Symposium on Operating Systems Design and
  Implementation (OSDI 23)}, pages 51--64, 2023.

\bibitem[Bershad et~al.(1995)Bershad, Savage, Pardyak, Sirer, Fiuczynski,
  Becker, Chambers, and Eggers]{Bershad95}
B.~N. Bershad, S.~Savage, P.~Pardyak, E.~G. Sirer, M.~E. Fiuczynski, D.~Becker,
  C.~Chambers, and S.~Eggers.
\newblock Extensibility safety and performance in the spin operating system.
\newblock In \emph{Proceedings of the Fifteenth ACM Symposium on Operating
  Systems Principles}, SOSP '95, page 267–283, New York, NY, USA, 1995.
  Association for Computing Machinery.
\newblock ISBN 0897917154.
\newblock \doi{10.1145/224056.224077}.
\newblock URL \url{https://doi.org/10.1145/224056.224077}.

\bibitem[Cadar et~al.(2008)Cadar, Dunbar, Engler, et~al.]{klee}
C.~Cadar, D.~Dunbar, D.~R. Engler, et~al.
\newblock Klee: Unassisted and automatic generation of high-coverage tests for
  complex systems programs.
\newblock In \emph{OSDI}, volume~8, pages 209--224, 2008.

\bibitem[Charalambous et~al.(2023)Charalambous, Tihanyi, Jain, Sun, Ferrag, and
  Cordeiro]{charalambous2023new}
Y.~Charalambous, N.~Tihanyi, R.~Jain, Y.~Sun, M.~A. Ferrag, and L.~C. Cordeiro.
\newblock A new era in software security: Towards self-healing software via
  large language models and formal verification.
\newblock \emph{arXiv preprint arXiv:2305.14752}, 2023.

\bibitem[Chen et~al.(2021{\natexlab{a}})Chen, Tworek, Jun, Yuan, Pinto, Kaplan,
  Edwards, Burda, Joseph, Brockman, et~al.]{chen21}
M.~Chen, J.~Tworek, H.~Jun, Q.~Yuan, H.~P. d.~O. Pinto, J.~Kaplan, H.~Edwards,
  Y.~Burda, N.~Joseph, G.~Brockman, et~al.
\newblock Evaluating large language models trained on code.
\newblock \emph{arXiv preprint arXiv:2107.03374}, 2021{\natexlab{a}}.

\bibitem[Chen et~al.(2021{\natexlab{b}})Chen, Tworek, Jun, Yuan, Pinto, Kaplan,
  Edwards, Burda, Joseph, Brockman, et~al.]{human-eval}
M.~Chen, J.~Tworek, H.~Jun, Q.~Yuan, H.~P. d.~O. Pinto, J.~Kaplan, H.~Edwards,
  Y.~Burda, N.~Joseph, G.~Brockman, et~al.
\newblock Evaluating large language models trained on code.
\newblock \emph{arXiv preprint arXiv:2107.03374}, 2021{\natexlab{b}}.

\bibitem[Chen et~al.(2023)Chen, Lin, Sch{\"a}rli, and Zhou]{chen2023teaching}
X.~Chen, M.~Lin, N.~Sch{\"a}rli, and D.~Zhou.
\newblock Teaching large language models to self-debug.
\newblock \emph{arXiv preprint arXiv:2304.05128}, 2023.

\bibitem[Chipounov et~al.(2009)Chipounov, Georgescu, Zamfir, and
  Candea]{chipounov2009selective}
V.~Chipounov, V.~Georgescu, C.~Zamfir, and G.~Candea.
\newblock Selective symbolic execution.
\newblock In \emph{Proceedings of the 5th Workshop on Hot Topics in System
  Dependability (HotDep)}, number CONF, 2009.

\bibitem[Chipounov et~al.(2011)Chipounov, Kuznetsov, and Candea]{Chipounov11}
V.~Chipounov, V.~Kuznetsov, and G.~Candea.
\newblock S2e: A platform for in-vivo multi-path analysis of software systems.
\newblock In \emph{Proceedings of the Sixteenth International Conference on
  Architectural Support for Programming Languages and Operating Systems},
  ASPLOS XVI, page 265–278, New York, NY, USA, 2011. Association for
  Computing Machinery.
\newblock ISBN 9781450302661.
\newblock \doi{10.1145/1950365.1950396}.
\newblock URL \url{https://doi.org/10.1145/1950365.1950396}.

\bibitem[Clarke and Emerson(1982)]{Clarke82}
E.~M. Clarke and E.~A. Emerson.
\newblock Design and synthesis of synchronization skeletons using branching
  time temporal logic.
\newblock In D.~Kozen, editor, \emph{Logics of Programs}, pages 52--71, Berlin,
  Heidelberg, 1982. Springer Berlin Heidelberg.
\newblock ISBN 978-3-540-39047-3.

\bibitem[Cloudflare(2023)]{ebpf-exporter}
Cloudflare.
\newblock ebpf\_exporter: ebpf-based exporter for prometheus.
\newblock GitHub repository, 2023.
\newblock \url{https://github.com/cloudflare/ebpf_exporter}.

\bibitem[Community(2023)]{alibaba23}
A.~C.~N. Community.
\newblock Seven core issues about ebpf, 2023.
\newblock
  \url{https://www.alibabacloud.com/blog/seven-core-issues-about-ebpf_599668}.

\bibitem[De~Moura and Bj{\o}rner(2008)]{DeMoura08}
L.~De~Moura and N.~Bj{\o}rner.
\newblock Z3: An efficient smt solver.
\newblock In \emph{International conference on Tools and Algorithms for the
  Construction and Analysis of Systems}, pages 337--340. Springer, 2008.

\bibitem[eBPF~for Windows~Contributors(2023)]{eBPFWindows}
eBPF~for Windows~Contributors.
\newblock ebpf for windows, 2023.
\newblock \url{https://github.com/microsoft/ebpf-for-windows}.

\bibitem[Elnaggar et~al.(2021)Elnaggar, Ding, Jones, Gibbs, Feher, Angerer,
  Severini, Matthes, and Rost]{Elnaggar21}
A.~Elnaggar, W.~Ding, L.~Jones, T.~Gibbs, T.~Feher, C.~Angerer, S.~Severini,
  F.~Matthes, and B.~Rost.
\newblock Codetrans: Towards cracking the language of silicon's code through
  self-supervised deep learning and high performance computing, 2021.

\bibitem[Feng et~al.(2020)Feng, Guo, Tang, Duan, Feng, Gong~(YIMING), Shou,
  Qin, Liu, Jiang, and Zhou]{feng20}
Z.~Feng, D.~Guo, D.~Tang, N.~Duan, X.~Feng, M.~Gong~(YIMING), L.~Shou, B.~Qin,
  T.~Liu, D.~Jiang, and M.~Zhou.
\newblock Codebert: A pre-trained model for programming and natural languages.
\newblock In \emph{Findings of EMNLP 2020}, September 2020.
\newblock URL
  \url{https://www.microsoft.com/en-us/research/publication/codebert-a-pre-trained-model-for-programming-and-natural-languages/}.

\bibitem[fuzzing~book author()]{fuzzing}
fuzzing~book author.
\newblock The fuzzing book: Concolic fuzzing.
\newblock \url{https://www.fuzzingbook.org/beta/html/SymbolicFuzzer.html}.

\bibitem[Ghigoff et~al.(2021)Ghigoff, Sopena, Lazri, Blin, and
  Muller]{ghigoff2021bmc}
Y.~Ghigoff, J.~Sopena, K.~Lazri, A.~Blin, and G.~Muller.
\newblock $\{$BMC$\}$: Accelerating memcached using safe in-kernel caching and
  pre-stack processing.
\newblock In \emph{18th USENIX Symposium on Networked Systems Design and
  Implementation (NSDI 21)}, pages 487--501, 2021.

\bibitem[Gregg(2001)]{Gregg01}
B.~Gregg.
\newblock Brenden gregg's homepage, 2001.
\newblock \url{https://www.brendangregg.com/}.

\bibitem[Gregg(2016)]{gregg16}
B.~Gregg.
\newblock Linux extended bpf (ebpf) tracing tools, 2016.
\newblock \url{https://www.brendangregg.com/ebpf.html}.

\bibitem[Gregg(2020)]{Gregg20}
B.~Gregg.
\newblock Bpf binaries: Btf, co-re, and the future of bpf perl tools, 2020.
\newblock
  \url{https://www.brendangregg.com/blog/2020-11-04/bpf-co-re-btf-libbpf.html}.

\bibitem[Gregg(2021)]{gregg2021computing}
B.~Gregg.
\newblock Computing performance.
\newblock 2021.

\bibitem[Gulwani et~al.(2011)Gulwani, Jha, Tiwari, and Venkatesan]{Gulwani11}
S.~Gulwani, S.~Jha, A.~Tiwari, and R.~Venkatesan.
\newblock Synthesis of loop-free programs.
\newblock In \emph{Proceedings of the 32nd ACM SIGPLAN Conference on
  Programming Language Design and Implementation}, PLDI '11, page 62–73, New
  York, NY, USA, 2011. Association for Computing Machinery.
\newblock ISBN 9781450306638.
\newblock \doi{10.1145/1993498.1993506}.
\newblock URL \url{https://doi.org/10.1145/1993498.1993506}.

\bibitem[Gurfinkel et~al.(2015)Gurfinkel, Kahsai, and
  Navas]{gurfinkel2015seahorn}
A.~Gurfinkel, T.~Kahsai, and J.~A. Navas.
\newblock Seahorn: A framework for verifying c programs (competition
  contribution).
\newblock In \emph{International Conference on Tools and Algorithms for the
  Construction and Analysis of Systems}, pages 447--450. Springer, 2015.

\bibitem[huangting4201()]{intern-lm}
e.~huangting4201, yingtongxiong.
\newblock Internlm: Chat models tailored for practical scenarios and the
  training system.
\newblock \url{https://github.com/InternLM/InternLM/}.

\bibitem[Humphries et~al.(2021)Humphries, Natu, Chaugule, Weisse, Rhoden, Don,
  Rizzo, Rombakh, Turner, and Kozyrakis]{humphries2021ghost}
J.~T. Humphries, N.~Natu, A.~Chaugule, O.~Weisse, B.~Rhoden, J.~Don, L.~Rizzo,
  O.~Rombakh, P.~Turner, and C.~Kozyrakis.
\newblock ghost: Fast \& flexible user-space delegation of linux scheduling.
\newblock In \emph{Proceedings of the ACM SIGOPS 28th Symposium on Operating
  Systems Principles}, pages 588--604, 2021.

\bibitem[Intellegence()]{langchain}
R.~Intellegence.
\newblock Langchain.
\newblock \url{https://www.langchain.com/}.

\bibitem[Jha et~al.(2010)Jha, Gulwani, Seshia, and Tiwari]{Jha10}
S.~Jha, S.~Gulwani, S.~A. Seshia, and A.~Tiwari.
\newblock Oracle-guided component-based program synthesis.
\newblock In \emph{Proceedings of the 32nd ACM/IEEE International Conference on
  Software Engineering - Volume 1}, ICSE '10, page 215–224, New York, NY,
  USA, 2010. Association for Computing Machinery.
\newblock ISBN 9781605587196.
\newblock \doi{10.1145/1806799.1806833}.
\newblock URL \url{https://doi.org/10.1145/1806799.1806833}.

\bibitem[Jia et~al.(2023{\natexlab{a}})Jia, Le, Ahmed, Williams, and
  Jamjoom]{Jia23_cfi}
J.~Jia, M.~V. Le, S.~Ahmed, D.~Williams, and H.~Jamjoom.
\newblock Practical and flexible kernel cfi enforcement using ebpf.
\newblock In \emph{Proceedings of the 1st Workshop on EBPF and Kernel
  Extensions}, eBPF '23, page 84–85, New York, NY, USA, 2023{\natexlab{a}}.
  Association for Computing Machinery.
\newblock ISBN 9798400702938.
\newblock \doi{10.1145/3609021.3609293}.
\newblock URL \url{https://doi.org/10.1145/3609021.3609293}.

\bibitem[Jia et~al.(2023{\natexlab{b}})Jia, Le, Ahmed, Williams, and
  Jamjoom]{jia2023practical}
J.~Jia, M.~V. Le, S.~Ahmed, D.~Williams, and H.~Jamjoom.
\newblock Practical and flexible kernel cfi enforcement using ebpf.
\newblock In \emph{Proceedings of the 1st Workshop on eBPF and Kernel
  Extensions}, pages 84--85, 2023{\natexlab{b}}.

\bibitem[Jia et~al.(2023{\natexlab{c}})Jia, Sahu, Oswald, Williams, Le, and
  Xu]{Jia23}
J.~Jia, R.~Sahu, A.~Oswald, D.~Williams, M.~V. Le, and T.~Xu.
\newblock f.
\newblock In \emph{Proceedings of the 19th Workshop on Hot Topics in Operating
  Systems}, pages 150--157, 2023{\natexlab{c}}.

\bibitem[Jia et~al.(2023{\natexlab{d}})Jia, Zhu, Williams, Arcangeli, Canella,
  Franke, Feldman-Fitzthum, Skarlatos, Gruss, and Xu]{jia2023programmable}
J.~Jia, Y.~Zhu, D.~Williams, A.~Arcangeli, C.~Canella, H.~Franke,
  T.~Feldman-Fitzthum, D.~Skarlatos, D.~Gruss, and T.~Xu.
\newblock Programmable system call security with ebpf.
\newblock \emph{arXiv preprint arXiv:2302.10366}, 2023{\natexlab{d}}.

\bibitem[Li et~al.(2022)Li, Choi, Chung, Kushman, Schrittwieser, Leblond,
  Eccles, Keeling, Gimeno, Lago, Hubert, Choy, de~Masson~d'Autume, Babuschkin,
  Chen, Huang, Welbl, Gowal, Cherepanov, Molloy, Mankowitz, Robson, Kohli,
  de~Freitas, Kavukcuoglu, and Vinyals]{Li22}
Y.~Li, D.~Choi, J.~Chung, N.~Kushman, J.~Schrittwieser, R.~Leblond, T.~Eccles,
  J.~Keeling, F.~Gimeno, A.~D. Lago, T.~Hubert, P.~Choy, C.~de~Masson~d'Autume,
  I.~Babuschkin, X.~Chen, P.-S. Huang, J.~Welbl, S.~Gowal, A.~Cherepanov,
  J.~Molloy, D.~J. Mankowitz, E.~S. Robson, P.~Kohli, N.~de~Freitas,
  K.~Kavukcuoglu, and O.~Vinyals.
\newblock Competition-level code generation with {AlphaCode}.
\newblock \emph{Science}, 378\penalty0 (6624):\penalty0 1092--1097, dec 2022.
\newblock \doi{10.1126/science.abq1158}.
\newblock URL \url{https://doi.org/10.1126%2Fscience.abq1158}.

\bibitem[Liu et~al.(2023)Liu, Xia, Wang, and Zhang]{liu2023your}
J.~Liu, C.~S. Xia, Y.~Wang, and L.~Zhang.
\newblock Is your code generated by chatgpt really correct? rigorous evaluation
  of large language models for code generation.
\newblock \emph{arXiv preprint arXiv:2305.01210}, 2023.

\bibitem[Luo et~al.(2023)Luo, Xu, Zhao, Sun, Geng, Hu, Tao, Ma, Lin, and
  Jiang]{luo2023wizardcoder}
Z.~Luo, C.~Xu, P.~Zhao, Q.~Sun, X.~Geng, W.~Hu, C.~Tao, J.~Ma, Q.~Lin, and
  D.~Jiang.
\newblock Wizardcoder: Empowering code large language models with
  evol-instruct.
\newblock \emph{arXiv preprint arXiv:2306.08568}, 2023.

\bibitem[Mayer et~al.(2021)Mayer, Loreti, Bracciale, Lungaroni, Salsano, and
  Filsfils]{Mayer21}
A.~Mayer, P.~Loreti, L.~Bracciale, P.~Lungaroni, S.~Salsano, and C.~Filsfils.
\newblock Performance monitoring with h2: Hybrid kernel/ebpf data plane for
  srv6 based hybrid sdn.
\newblock \emph{Computer Networks}, 185:\penalty0 107705, 2021.
\newblock ISSN 1389-1286.
\newblock \doi{https://doi.org/10.1016/j.comnet.2020.107705}.
\newblock URL
  \url{https://www.sciencedirect.com/science/article/pii/S1389128620313037}.

\bibitem[Necula and Lee(1996)]{Necula96}
G.~C. Necula and P.~Lee.
\newblock Safe kernel extensions without run-time checking.
\newblock In \emph{Proceedings of the Second USENIX Symposium on Operating
  Systems Design and Implementation}, OSDI '96, page 229–243, New York, NY,
  USA, 1996. Association for Computing Machinery.
\newblock ISBN 1880446820.
\newblock \doi{10.1145/238721.238781}.
\newblock URL \url{https://doi.org/10.1145/238721.238781}.

\bibitem[OpenAI()]{instruct-gpt}
OpenAI.
\newblock Instructgpt: Aligning language models to follow instructions.
\newblock
  \url{https://openai.com/research/instruction-following#ref-ASamir%20Rajadnya}.

\bibitem[Perkins and Ernst(2004)]{Perkins04}
J.~H. Perkins and M.~D. Ernst.
\newblock Efficient incremental algorithms for dynamic detection of likely
  invariants.
\newblock In \emph{Proceedings of the 12th ACM SIGSOFT Twelfth International
  Symposium on Foundations of Software Engineering}, SIGSOFT '04/FSE-12, page
  23–32, New York, NY, USA, 2004. Association for Computing Machinery.
\newblock ISBN 1581138555.
\newblock \doi{10.1145/1029894.1029901}.
\newblock URL \url{https://doi.org/10.1145/1029894.1029901}.

\bibitem[Phan et~al.(2021)Phan, Tran, Le, Nguyen, Anibal, Peltekian, and
  Ye]{Phan21}
L.~Phan, H.~Tran, D.~Le, H.~Nguyen, J.~Anibal, A.~Peltekian, and Y.~Ye.
\newblock Cotext: Multi-task learning with code-text transformer.
\newblock \emph{arXiv preprint arXiv:2105.08645}, 2021.

\bibitem[Poeplau and Francillon(2020)]{poeplau2020symbolic}
S.~Poeplau and A.~Francillon.
\newblock Symbolic execution with $\{$SymCC$\}$: Don't interpret, compile!
\newblock In \emph{29th USENIX Security Symposium (USENIX Security 20)}, pages
  181--198, 2020.

\bibitem[Poesia et~al.(2023)Poesia, Gandhi, Zelikman, and
  Goodman]{poesia2023certified}
G.~Poesia, K.~Gandhi, E.~Zelikman, and N.~D. Goodman.
\newblock Certified reasoning with language models.
\newblock \emph{arXiv preprint arXiv:2306.04031}, 2023.

\bibitem[Project(2023)]{bcc}
I.~V. Project.
\newblock Bpf compiler collection (bcc), 2023.
\newblock Available: \url{https://github.com/iovisor/bcc}.

\bibitem[Queille and Sifakis(1982)]{Queille82}
J.~P. Queille and J.~Sifakis.
\newblock Specification and verification of concurrent systems in cesar.
\newblock In M.~Dezani-Ciancaglini and U.~Montanari, editors,
  \emph{International Symposium on Programming}, pages 337--351, Berlin,
  Heidelberg, 1982. Springer Berlin Heidelberg.
\newblock ISBN 978-3-540-39184-5.

\bibitem[Rozi{\`e}re et~al.(2023)Rozi{\`e}re, Gehring, Gloeckle, Sootla, Gat,
  Tan, Adi, Liu, Remez, Rapin, et~al.]{roziere2023code}
B.~Rozi{\`e}re, J.~Gehring, F.~Gloeckle, S.~Sootla, I.~Gat, X.~E. Tan, Y.~Adi,
  J.~Liu, T.~Remez, J.~Rapin, et~al.
\newblock Code llama: Open foundation models for code.
\newblock \emph{arXiv preprint arXiv:2308.12950}, 2023.

\bibitem[Shah()]{tpu}
A.~Shah.
\newblock Google tpu v5e ai chip debuts after controversial origins.
\newblock
  \url{https://www.enterpriseai.news/2023/08/31/google-tpu-v5e-ai-chip-debuts-after-controversial-origins/}.

\bibitem[Solar-Lezama et~al.(2006)Solar-Lezama, Tancau, Bodik, Seshia, and
  Saraswat]{SolarLezama06}
A.~Solar-Lezama, L.~Tancau, R.~Bodik, S.~Seshia, and V.~Saraswat.
\newblock Combinatorial sketching for finite programs.
\newblock In \emph{Proceedings of the 12th International Conference on
  Architectural Support for Programming Languages and Operating Systems},
  ASPLOS XII, page 404–415, New York, NY, USA, 2006. Association for
  Computing Machinery.
\newblock ISBN 1595934510.
\newblock \doi{10.1145/1168857.1168907}.
\newblock URL \url{https://doi.org/10.1145/1168857.1168907}.

\bibitem[Sun et~al.(2021)Sun, Shen, Wang, Liu, Jiang, Chen, and
  Cui]{sun2021healer}
H.~Sun, Y.~Shen, C.~Wang, J.~Liu, Y.~Jiang, T.~Chen, and A.~Cui.
\newblock Healer: Relation learning guided kernel fuzzing.
\newblock In \emph{Proceedings of the ACM SIGOPS 28th Symposium on Operating
  Systems Principles}, pages 344--358, 2021.

\bibitem[Swift et~al.(2003)Swift, Bershad, and Levy]{Swift03}
M.~M. Swift, B.~N. Bershad, and H.~M. Levy.
\newblock Improving the reliability of commodity operating systems.
\newblock In \emph{Proceedings of the Nineteenth ACM Symposium on Operating
  Systems Principles}, SOSP '03, page 207–222, New York, NY, USA, 2003.
  Association for Computing Machinery.
\newblock ISBN 1581137575.
\newblock \doi{10.1145/945445.945466}.
\newblock URL \url{https://doi.org/10.1145/945445.945466}.

\bibitem[team()]{chat-gpt}
O.~team.
\newblock Internlm: Chat models tailored for practical scenarios and the
  training system.
\newblock \url{https://chat.openai.com/}.

\bibitem[Team({\natexlab{a}})]{pixel6}
T.~G.~P. Team.
\newblock The source code of google pixel6 android kernel, {\natexlab{a}}.
\newblock \url{https://android.googlesource.com/device/google/raviole-kernel/}.

\bibitem[Team({\natexlab{b}})]{lwn-verifier}
T.~L. Team.
\newblock The ebpf verifier, {\natexlab{b}}.
\newblock \url{https://static.lwn.net/kerneldoc/bpf/verifier.html}.

\bibitem[Torlak and Bodik(2013)]{Torlak13}
E.~Torlak and R.~Bodik.
\newblock Growing solver-aided languages with rosette.
\newblock In \emph{Proceedings of the 2013 ACM International Symposium on New
  Ideas, New Paradigms, and Reflections on Programming \& Software}, Onward!
  2013, page 135–152, New York, NY, USA, 2013. Association for Computing
  Machinery.
\newblock ISBN 9781450324724.
\newblock \doi{10.1145/2509578.2509586}.
\newblock URL \url{https://doi.org/10.1145/2509578.2509586}.

\bibitem[Visor(2023)]{bpftrace}
I.~Visor.
\newblock bpftrace: High-level tracing language for linux ebpf.
\newblock GitHub repository, 2023.
\newblock \url{https://github.com/iovisor/bpftrace}.

\bibitem[Wang et~al.(2021)Wang, Yi, Guo, Jin, Xu, Li, Wang, Guo, Li, Xu,
  et~al.]{Wang21}
J.~Wang, X.~Yi, R.~Guo, H.~Jin, P.~Xu, S.~Li, X.~Wang, X.~Guo, C.~Li, X.~Xu,
  et~al.
\newblock Milvus: A purpose-built vector data management system.
\newblock In \emph{Proceedings of the 2021 International Conference on
  Management of Data}, pages 2614--2627, 2021.

\bibitem[Wang et~al.(2020)Wang, Yao, Kwok, and Ni]{Wang20}
Y.~Wang, Q.~Yao, J.~T. Kwok, and L.~M. Ni.
\newblock Generalizing from a few examples: A survey on few-shot learning.
\newblock \emph{ACM computing surveys (csur)}, 53\penalty0 (3):\penalty0 1--34,
  2020.

\bibitem[Wikipedia()]{harmony-wiki}
Wikipedia.
\newblock The wikipedia of harmonyos.
\newblock \url{https://en.wikipedia.org/wiki/HarmonyOS}.

\bibitem[Wu et~al.(2020)Wu, Zhao, and Zhang]{Wu20}
H.~Wu, H.~Zhao, and M.~Zhang.
\newblock Code summarization with structure-induced transformer.
\newblock \emph{arXiv preprint arXiv:2012.14710}, 2020.

\bibitem[Xu et~al.(2021)Xu, Wong, Wagle, Narayana, and Sivaraman]{Xu21}
Q.~Xu, M.~D. Wong, T.~Wagle, S.~Narayana, and A.~Sivaraman.
\newblock Synthesizing safe and efficient kernel extensions for packet
  processing.
\newblock In \emph{Proceedings of the 2021 ACM SIGCOMM 2021 Conference},
  SIGCOMM '21, page 50–64, New York, NY, USA, 2021. Association for Computing
  Machinery.
\newblock ISBN 9781450383837.
\newblock \doi{10.1145/3452296.3472929}.
\newblock URL \url{https://doi.org/10.1145/3452296.3472929}.

\bibitem[Yang et~al.(2023)Yang, Lu, Liao, Chen, Li, He, and
  Shu]{yang2023lambda}
Z.~Yang, Y.~Lu, X.~Liao, Y.~Chen, J.~Li, S.~He, and J.~Shu.
\newblock $\{$$\lambda$-IO$\}$: A unified $\{$IO$\}$ stack for computational
  storage.
\newblock In \emph{21st USENIX Conference on File and Storage Technologies
  (FAST 23)}, pages 347--362, 2023.

\bibitem[Zhong et~al.(2022)Zhong, Li, Wu, Zarkadas, Tao, Mesterhazy, Makris,
  Yang, Tai, Stutsman, and Cidon]{Zhong22}
Y.~Zhong, H.~Li, Y.~J. Wu, I.~Zarkadas, J.~Tao, E.~Mesterhazy, M.~Makris,
  J.~Yang, A.~Tai, R.~Stutsman, and A.~Cidon.
\newblock {XRP}: {In-Kernel} storage functions with {eBPF}.
\newblock In \emph{16th USENIX Symposium on Operating Systems Design and
  Implementation (OSDI 22)}, pages 375--393, Carlsbad, CA, July 2022. USENIX
  Association.
\newblock ISBN 978-1-939133-28-1.
\newblock URL
  \url{https://www.usenix.org/conference/osdi22/presentation/zhong}.

\bibitem[Z{\"u}gner et~al.(2021)Z{\"u}gner, Kirschstein, Catasta, Leskovec, and
  G{\"u}nnemann]{Zugner21}
D.~Z{\"u}gner, T.~Kirschstein, M.~Catasta, J.~Leskovec, and S.~G{\"u}nnemann.
\newblock Language-agnostic representation learning of source code from
  structure and context.
\newblock \emph{arXiv preprint arXiv:2103.11318}, 2021.

\end{thebibliography}

\end{document}